\documentclass{article} 
\usepackage{iclr2027_conference,times}

\usepackage{amsmath,amsfonts,bm}

\def\eqref#1{equation~\ref{#1}}

\def\1{\bm{1}}

\DeclareMathAlphabet{\mathsfit}{\encodingdefault}{\sfdefault}{m}{sl}
\SetMathAlphabet{\mathsfit}{bold}{\encodingdefault}{\sfdefault}{bx}{n}

\DeclareMathOperator*{\argmax}{arg\,max}

\usepackage{hyperref}
\usepackage{url}
\usepackage[table]{xcolor}
\usepackage{array}
\usepackage{microtype}
\usepackage{graphicx}
\usepackage{tcolorbox}
\tcbuselibrary{breakable,skins,raster}
\usepackage{xcolor,pifont,booktabs}
\usepackage{subcaption}
\usepackage{booktabs}
\usepackage{amsmath}
\usepackage{amssymb}
\usepackage{mathtools}
\usepackage{amsthm}
\usepackage{times}
\usepackage{latexsym}
\usepackage[T1]{fontenc}
\usepackage[utf8]{inputenc}
\usepackage{graphicx}
\usepackage{xspace}
\usepackage{tabularx}
\usepackage{wrapfig}
\usepackage{multirow}
\usepackage{graphicx}
\usepackage{pifont}
\usepackage{float}
\usepackage{enumitem}
\usepackage{times}
\usepackage{latexsym}
\usepackage{booktabs}
\usepackage{arydshln}
\usepackage{makecell}
\usepackage{comment}
\usepackage{cleveref}
\usepackage{placeins}
\usepackage{stfloats}
\usepackage{dashrule}
\usepackage[flushmargin]{footmisc}

\title{Rubric-Aware On-Policy Self-Distillation for LLM Personalization}

\author{\parbox[t]{0.97\textwidth}{\raggedright
\vspace{-12pt}
\bfseries
Yilun Qiu\raisebox{0.6ex}{\scriptsize 1,2}\quad
Xiaoyan Zhao\raisebox{0.6ex}{\scriptsize 2\ding{41}}\quad
Chengbing Wang\raisebox{0.6ex}{\scriptsize 3}\quad
Cilin Yan\raisebox{0.6ex}{\scriptsize 1}\quad
Rui Zu\raisebox{0.6ex}{\scriptsize 1,4}\\[3pt]
Wanyang Zhang\raisebox{0.6ex}{\scriptsize 1,4}\quad
Xiaolong Jiang\raisebox{0.6ex}{\scriptsize 1}\quad
Jiayin Cai\raisebox{0.6ex}{\scriptsize 1}\quad
Yang Zhang\raisebox{0.6ex}{\scriptsize 2}\\[5pt]
\normalfont
Xiaohongshu Inc.\raisebox{0.6ex}{\scriptsize 1}\quad
National University of Singapore\raisebox{0.6ex}{\scriptsize 2}\\[2pt]
University of Science and Technology of China\raisebox{0.6ex}{\scriptsize 3}\quad
Peking University\raisebox{0.6ex}{\scriptsize 4}\\[3pt]
{\small\ttfamily \ding{41} Corresponding author}\\[5pt]
{\small\hypersetup{pdfborder={0 0 0}}\href{https://github.com/SnowCharmQ/GRASP}{\textcolor{magenta}{\texttt{https://github.com/SnowCharmQ/GRASP}}}}
}}

\newcommand{\cmark}{\textcolor{green}{\checkmark}}
\newcommand{\xmark}{\textcolor{red}{\ding{55}}}
\newcolumntype{C}{>{\centering\arraybackslash}X}

\newcommand{\ours}{GRASP\xspace}
\definecolor{oursgray}{gray}{0.96}

\iclrfinalcopy 
\begin{document}

\maketitle
\fancyhead[L]{\small Rubric-Aware On-Policy Self-Distillation for LLM Personalization}

\begin{abstract}
LLM personalization aims to generate responses aligned with individual users' preferences and needs.
User-specific rubrics make these expectations explicit, providing direct supervision on what a satisfactory answer should cover.
Existing rubric-guided approaches, however, exploit such guidance only at a coarse granularity, either by using rubrics to supervise the prediction of relevant aspects for subsequent generation or by reducing aspect coverage to a single response-level reward for reinforcement learning.
This leaves a gap between specifying what a personalized answer should contain and teaching the model how to generate it.
To bridge this gap, we propose \textbf{\ours}, a rubric-aware on-policy self-distillation framework for LLM personalization that turns user-specific rubric aspects into fine-grained, token-level supervision.
Specifically, \ours pairs a rubric-free student with a rubric-informed teacher that additionally receives the target user-specific rubrics.
By aligning their next-token distributions along on-policy trajectories generated by the student, \ours transfers the teacher's rubric-conditioned guidance into the student, translating user-specific semantic requirements into dense token-level supervision.
Since rubric-informed teachers can still produce inadequate supervision, we further introduce Rubric-based Teacher Validation~(RTV), which retains only instances where the teacher sufficiently covers the target aspects, improving both supervision quality and training efficiency.
Experiments on the LaMP-QA benchmark for personalized question answering demonstrate that \ours achieves state-of-the-art performance across multiple backbones, supporting the effectiveness of rubric-guided token-level supervision for personalization.

\end{abstract}

\section{Introduction}

As large language models (LLMs)~\citep{gpt4,deepseekr1,qwen3,deepseekv4} become increasingly capable, their expanding use in information seeking and decision support has heightened the need to serve users with diverse preferences and needs~\citep{lamp,nextquill,zhang2026reinforced}.
This growing use motivates \emph{LLM personalization}, which moves beyond ``one-size-fits-all'' responses to accommodate differences among users~\citep{personalizationsurvey2,personalizationsurvey4,dpl}.
This need is particularly salient in question answering, where users asking the same question may seek different information, motivating \emph{personalized question answering}~(PQA)~\citep{lampqa}.

What makes a personalized answer good is not merely factual correctness, but whether it covers the aspects that matter to this particular user~\citep{parl,vitabench2}.
Two users asking the same question may prioritize entirely different things: one may favor conventional choices, whereas another may seek novelty, so an answer that is accurate yet addresses the wrong aspects helps neither.
These expectations are increasingly formalized as \emph{user-specific rubrics}, whose constituent aspects specify what a satisfactory personalized answer should cover~\citep{lampqa}.
The central challenge is therefore to translate these rubric-defined requirements into effective guidance for generating answers that address the user's specific needs.

Yet existing rubric-guided personalization methods offer limited guidance on how to realize these requirements during generation.
PlanPers~\citep{lampqa} trains a separate planner to predict relevant rubric aspects and conditions answer generation on the resulting plan, but rubric supervision targets the planner rather than the generator, and planning errors can affect the final response.
IAP~\citep{iap} instead converts aspect coverage into a scalar reward and optimizes it with reinforcement learning, but such a reward is sparse, providing limited feedback on which token-level decisions contribute to satisfying individual aspects.
Consequently, existing methods exploit rich, user-specific rubric aspects only through coarse supervision, leaving their potential for fine-grained token-level guidance largely underused.

On-policy self-distillation~(OPSD) offers a natural way to bridge this gap: a rubric-informed teacher can provide next-token supervision along the student's own generation trajectories, allowing a rubric-free student to learn from guidance unavailable at inference time~\citep{opsd}.
Motivated by this, we propose \textbf{\ours}, a \emph{Rubric-Aware On-Policy Self-Distillation} framework for LLM personalization.
Specifically, \ours initializes a trainable rubric-free student, while the same frozen model conditioned on the user-specific rubric aspects serves as the teacher.
For responses sampled from the student, both models predict the next token at each student-generated prefix.
The student minimizes the forward KL divergence between the teacher's and its own next-token distributions, receiving dense, rubric-conditioned supervision at the generation states it actually encounters.
By restricting rubric access to the teacher during training, \ours distills this privileged guidance into the student, enabling personalized generation without rubrics at inference time.
In this way, \ours translates user-specific rubric aspects into fine-grained token-level learning signals, enabling substantially finer credit assignment throughout generation.

Taking a step further, even when conditioned on the target rubric aspects, the teacher may still generate signals that inadequately reflect or cover them.
Distilling such underperforming teacher signals can steer the student in the wrong direction, and these uninformative rollouts also reduce training efficiency.
To address this, we introduce \emph{Rubric-based Teacher Validation}~(RTV), which validates each teacher output against the target rubrics and retains only those where the rubric-informed teacher demonstrably provides adequate aspect coverage.
RTV therefore improves both the reliability of distillation and sample efficiency by focusing learning on informative teacher signals.

We evaluate \ours on LaMP-QA~\citep{lampqa}, a large-scale PQA benchmark spanning three diverse domains.
Experimental results show that \ours achieves state-of-the-art performance across different backbone models.

The main contributions of this work are summarized as follows:
\begin{itemize}[leftmargin=*, topsep=0pt, itemsep=0pt]
    \item We identify a key limitation of existing rubric-guided personalization approaches: they provide limited guidance on how to realize user-specific rubric aspects during generation, leaving the potential of rubrics for fine-grained, token-level supervision largely underused.
    \item To the best of our knowledge, we are the first to bring on-policy self-distillation to personalization, where a rubric-informed teacher transfers the guidance of user-specific rubric aspects to the student through token-level supervision.
    \item We propose \ours, a Rubric-Aware On-Policy Self-Distillation framework for personalization, equipped with Rubric-based Teacher Validation that filters unreliable teacher signals to improve supervision quality and training efficiency.
    \item Extensive experiments on personalized question answering using LaMP-QA across multiple backbones demonstrate that \ours consistently achieves state-of-the-art performance.
\end{itemize}
\section{Related Work}

\par
\textbf{LLM Personalization.}
Personalization is essential for LLMs to serve each individual's distinct needs~\citep{personalizationsurvey2,personalizationsurvey4,personalizationsurvey3,personalizationsurvey8,luo2026know}, and has been pursued through training-free conditioning~\citep{pag,dpl}, supervised fine-tuning~\citep{oppu,nextquill}, and reinforcement learning~\citep{prlm,iap}.
Most of these efforts, however, target personalized content generation, where the model writes reviews, emails, or articles on a user's behalf~\citep{longlamp,hydra,dep,flythinker}, while far less attention has been paid to the case where users seek information rather than delegate writing: in personalized question answering, the model must infer \emph{what} a user needs to know rather than imitate \emph{how} they write.
LaMP-QA~\citep{lampqa} was introduced as a benchmark for this setting, pairing each question with a user profile and a set of user-specific rubric aspects that specify the information needs a satisfactory answer should address.
Building on it, IAP~\citep{iap} prompts the model to articulate the latent intent behind a question before answering and optimizes the resulting trajectories with RL against a reward derived from these aspects, while VAC~\citep{vac} replaces such scalar rewards with natural language critiques from a learned feedback model and fine-tunes the policy on the revised responses.
Across these paradigms, on-policy distillation remains unexplored for personalization, and our work aims to fill this gap.

\par
\textbf{On-Policy Self-Distillation.}
On-policy distillation (OPD)~\citep{opd} is a post-training paradigm that trains a student on trajectories sampled from its own policy, querying a teacher at each visited state for dense token-level supervision~\citep{gkd,opdsurvey,rethinkingopd}.
Recent studies have extensively explored its optimization strategies and applications, demonstrating strong potential across diverse reasoning and generation tasks~\citep{eopd,ropd,videoopd}.
Within this paradigm, on-policy self-distillation (OPSD)~\citep{opsd} instantiates the teacher and student from a single model conditioned on different contexts: only the teacher observes privileged information that is unavailable at deployment~\citep{skillsd,opsdl,lopd}.
Existing work has explored diverse forms of such context, including verified reasoning traces~\citep{opsd}, environment feedback~\citep{resd}, and evidence-guided context~\citep{edgeopd}.
Such contexts are externally verifiable or globally fixed; however, personalized QA provides user-specific rubric aspects: per-question semantic specifications.
To the best of our knowledge, we are the first to bring OPSD to LLM personalization.
\section{Preliminary}

In this section, we first formulate personalized question answering and then introduce on-policy self-distillation, which forms the foundation of our method.

\subsection{Problem Formulation}

This work studies personalized question answering~(PQA), which aims to generate responses tailored to the information needs of individual users.
For each user $u$, an instance consists of a question $x_u$ and a user profile $P_u$, which contains the user's historical interactions and provides evidence about their preferences and interests.
The goal is to generate a personalized response:
\begin{equation}
    y \sim \pi_{\theta}(\cdot \mid x_u, P_u),
\end{equation}
where $\pi_{\theta}$ denotes the language model parameterized by $\theta$.

Unlike conventional question answering, a satisfactory personalized response should not only be factually correct but also cover the information that matters to the individual user.
Following LaMP-QA~\citep{lampqa}, these user-specific information needs are represented by a set of rubric aspects, denoted as $E_{x_u} = \{e_1, e_2, \ldots, e_m\}$, where each aspect $e_i$ describes a criterion that a desirable response to $x_u$ should address for user $u$.
Given a response $y$, its personalization quality can therefore be characterized by its coverage of these aspects:
\begin{equation}
    R(y, E_{x_u})
    =
    \frac{1}{|E_{x_u}|}
    \sum_{e \in E_{x_u}} r(y,e),
\end{equation}
where $r(y,e)$ measures how well $y$ addresses aspect $e$.

During training, we are given a dataset
$\mathcal{D}=
    \left\{
        (x_u,P_u,E_{x_u})
    \right\}
$,
where each instance contains a question, a user profile, and a set of user-specific rubric aspects, but no gold reference response.
The objective of PQA is to maximize the expected rubric-based score of responses generated from the question and user profile:
\begin{equation}
    \theta^{*}
    =
    \argmax_{\theta}
    \mathbb{E}_{(x_u,P_u,E_{x_u})\sim\mathcal{D}}
    \mathbb{E}_{y\sim\pi_{\theta}(\cdot\mid x_u,P_u)}
    \left[
        R(y,E_{x_u})
    \right].
\end{equation}
At inference time, the model receives only $(x_u,P_u)$ and generates responses without access to $E_{x_u}$.

\subsection{On-Policy Self-Distillation}

On-policy distillation (OPD) trains a student on responses sampled from its own policy while using a teacher to provide dense token-level supervision~\citep{opd}.
On-policy self-distillation (OPSD) further instantiates the student and teacher from the same language model under different conditioning contexts~\citep{opsd}.
Given an input context $c$ and privileged information $z$, the student and teacher are defined as
$p_S(\cdot\mid c)=p_\theta(\cdot\mid c)$ and
$p_T(\cdot\mid c,z)=p_\theta(\cdot\mid c,z)$, respectively.
The OPSD learning objective is to minimize the per-token divergence $D(p_T\|p_S)$ between the teacher and student distributions along the student-generated response $\hat{y}$, denoted as:
\begin{equation}
    \mathcal{L}_{\mathrm{OPSD}}(\theta)
    =
    \mathbb{E}_{(c,z)\sim\mathcal{D}}
    \mathbb{E}_{\hat{y}\sim p_S(\cdot\mid c)}
    \left[
        D(p_T\|p_S)(\hat{y}\mid c,z)
    \right].
\end{equation}

\section{Methodology}\label{sec:method}

\begin{figure*}[!t]
    \centering
    \includegraphics[width=0.98\linewidth]{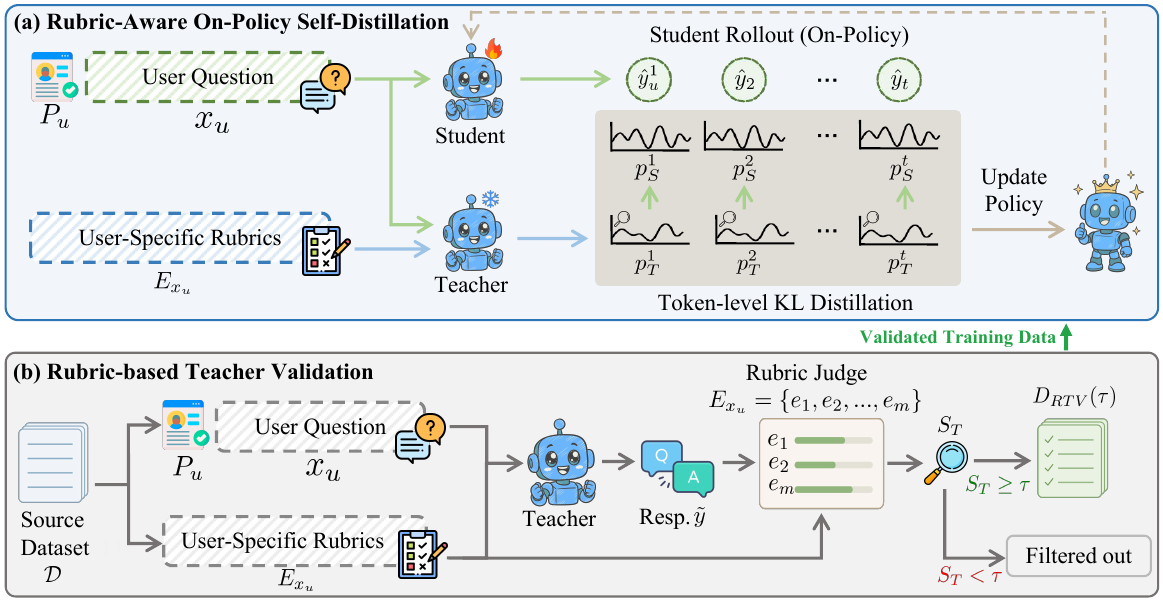}
    \vspace{-1mm}
    \caption{
Overview of our proposed \textbf{\ours} method.
(a) \emph{Rubric-Aware On-Policy Self-Distillation}: The student model learns from rubric-aware teacher distributions along its own on-policy trajectories.
(b) \emph{Rubric-based Teacher Validation}: This retains training instances whose teacher responses sufficiently cover the target rubric aspects.
}
    \label{main_method}
    \vspace{-3mm}
\end{figure*}

Personalized question answering requires generating responses that cover the aspects that matter to each individual user.
To this end, we propose \textbf{\ours}, a new training paradigm that brings on-policy self-distillation to PQA.
At its core, Rubric-Aware On-Policy Self-Distillation treats user-specific rubric aspects as privileged information and transfers their guidance to the student through dense token-level supervision.
To improve teacher reliability and training efficiency, we further introduce Rubric-based Teacher Validation, which retains only verified teacher signals for distillation.
Figure~\ref{main_method} provides an overview of \ours.
Next, we elaborate these components in detail.

\subsection{Rubric-Aware On-Policy Self-Distillation}

The core of \ours is to transform user-specific rubric aspects into dense supervision for PQA.
To achieve this, following the OPSD paradigm, for each training instance $(x_u,P_u,E_{x_u})\sim\mathcal{D}^*$ from the training dataset $\mathcal{D}^*$, we set the standard context as $c=(x_u,P_u)$ and the privileged information as $z=E_{x_u}$.

\par
\textbf{Rubric-free Student and Rubric-informed Teacher.}
Under this formulation, we instantiate a trainable rubric-free student and a frozen rubric-informed teacher from the same language model, with parameters $\theta$ and $\bar{\theta}$, respectively.
Their conditional policies are defined as:
\begin{equation}
    p_S(\cdot\mid c)
    \triangleq
    \pi_\theta(\cdot\mid c),
    \qquad
    p_T(\cdot\mid c,z)
    \triangleq
    \pi_{\bar{\theta}}(\cdot\mid c,z).
\end{equation}
The two policies share the same model architecture and vocabulary but differ in their conditioning contexts.
The student observes only the question and user profile, corresponding to the information available at inference time.
In contrast, the teacher additionally observes the target rubric aspects, which explicitly specify the user-specific information that the response should address.
Throughout training, the teacher parameters $\bar{\theta}$ remain fixed, while only the student parameters $\theta$ are optimized.

\par
\textbf{On-Policy Response Sampling.}
For each training instance, an on-policy response is sampled from the student policy:
\begin{equation}
    \hat{y}_u
    =
    (\hat{y}_u^1,\ldots,\hat{y}_u^{|\hat{y}_u|})
    \sim
    p_S(\cdot\mid c).
\end{equation}
Since the student conditions exclusively on $c=(x_u,P_u)$, the sampled response is generated under the same information setting used at inference time.
Distillation over these student-induced states therefore aligns the training state distribution with the inference state distribution.

\par
\textbf{Rubric-Aware Self-Distillation.}
Rather than generating a separate trajectory, the teacher evaluates the student-generated response under the privileged context $(c,z)$.
At each decoding step $t$, both policies predict the next token from the same student-generated prefix $\hat{y}_u^{1:t-1}$:
\begin{equation}
    p_S^t(\cdot)
    \triangleq
    p_S(\cdot\mid c,\hat{y}_u^{1:t-1}),
    \qquad
    p_T^t(\cdot)
    \triangleq
    p_T(\cdot\mid c,z,\hat{y}_u^{1:t-1}).
\end{equation}
Conditioning both policies on an identical prefix makes their next-token distributions directly comparable.
While the student distribution represents its current prediction based only on the question and user profile, the teacher distribution provides rubric-informed guidance that reflects which token predictions better address the user-specific aspects.

To transfer this guidance to the student, we align the teacher and student next-token distributions over the full vocabulary at each decoding step.
Formally, for a response $\hat{y}_u$ sampled from the student policy, we define the trajectory-averaged full-vocabulary divergence as:
\begin{equation}
    D(p_T\|p_S)(\hat{y}_u\mid c,z)
    \triangleq
    \frac{1}{|\hat{y}_u|}
    \sum_{t=1}^{|\hat{y}_u|}
    d\left(p_T^t\|p_S^t\right),
\end{equation}
where $d(\cdot\|\cdot)$ measures the divergence between two next-token distributions over the vocabulary $\mathcal{V}$.
In this work, we instantiate $d$ as the forward KL divergence:
\begin{equation}
    d\left(p_T^t\|p_S^t\right)
    =
    D_{\mathrm{KL}}\left(p_T^t\|p_S^t\right)
    =
    \sum_{v\in\mathcal{V}}
    p_T^t(v)
    \log\frac{p_T^t(v)}{p_S^t(v)}.
\end{equation}
Minimizing this divergence aligns the student with the rubric-informed teacher distribution, transferring fine-grained user-specific preferences over token predictions at each decoding step.

\subsection{Rubric-based Teacher Validation}

Although the rubric-informed teacher has access to the target aspects, privileged information alone does not guarantee reliable supervision.
The teacher may still overlook or insufficiently address some aspects when generating a response.
Since the rubric-aware self-distillation framework directly transfers the teacher's token distributions to the student, such failures may introduce misleading supervision.
To address this, we introduce Rubric-based Teacher Validation~(RTV), which evaluates the teacher's ability to utilize the provided rubric aspects and excludes unreliable instances before distillation.

For each training instance, we first generate a response from the frozen rubric-informed teacher:
\begin{equation}
    \tilde{y}_u
    \sim
    p_T(\cdot\mid c,z),
\end{equation}
where $\tilde{y}_u$ is used only for teacher validation.
We then employ a rubric judge to evaluate $\tilde{y}_u$ and define the teacher reliability score $s_T$ using the rubric-based scoring function:
\begin{equation}
    s_T
    \triangleq
    R(\tilde{y}_u,E_{x_u})
    =
    \frac{1}{|E_{x_u}|}
    \sum_{e\in E_{x_u}}
    r(\tilde{y}_u,e).
\end{equation}
Since $\tilde{y}_u$ is generated by the same rubric-informed policy used for distillation, its aspect coverage provides an observable measure of whether the teacher effectively utilizes the privileged rubric information.
A higher $s_T$ therefore indicates more reliable teacher guidance.

Based on the teacher reliability score, we construct the validated training set by retaining only instances that meet a threshold $\tau$:
\begin{equation}
    \mathcal{D}_{\mathrm{RTV}}(\tau)
    \triangleq
    \left\{
        (x_u,P_u,E_{x_u})\in\mathcal{D}
        \;\middle|\;
        s_T\geq\tau
    \right\}.
\end{equation}
Specifically, RTV is applied once to the original training set before distillation.
In our main setting, we set $\tau=1$, requiring the teacher response to fully address all target rubric aspects.
This strict criterion filters out unreliable instances, reducing the risk of transferring misleading rubric guidance to the student.

\subsection{Training Objective}
Combining Rubric-Aware On-Policy Self-Distillation with Rubric-based Teacher Validation, the overall training objective of \ours is to minimize the expected teacher--student divergence over the validated training dataset $\mathcal{D}_{\mathrm{RTV}}(\tau)$ and responses $\hat{y}_u$ sampled on-policy from $p_S(\cdot\mid c)$, formally:
\begin{equation}
    \mathcal{L}_{\text{\ours}}(\theta)
    =
    \mathbb{E}_{(x_u,P_u,E_{x_u})\sim{\mathcal{D}_{\mathrm{RTV}}(\tau)}}
    \mathbb{E}_{\hat{y}_u\sim p_S(\cdot\mid c)}
    \left[
        D(p_T\|p_S)(\hat{y}_u\mid c,z)
    \right].
\end{equation}
The teacher distributions are treated as fixed targets, with gradients propagated exclusively through the student policy.
Through token-level distribution alignment along on-policy trajectories, the student internalizes the guidance encoded by the rubric aspects while preserving the rubric-free information setting required at inference time.

\section{Experiments}

\subsection{Experimental Setup}\label{sec:experimental_setup}

\par
\textbf{Datasets \& Evaluation.}
We experiment on LaMP-QA~\citep{lampqa}, a comprehensive benchmark for personalized long-form question answering covering three main categories: \emph{Arts \& Entertainment}~(A\&E), \emph{Lifestyle \& Personal Development}~(L\&PD), and \emph{Society \& Culture}~(S\&C).
Each instance pairs a question $x_u$ with a user profile $P_u$ containing previously asked questions and a set of personalized rubric aspects $E_{x_u}$.
These aspects are derived from the question narrative $r_{x_u}$ and specify what a satisfactory personalized answer should address.
Following the benchmark protocol, each response is scored against every aspect $e \in E_{x_u}$ on a $0$–$2$ scale by an LLM judge, with normalized scores averaged to obtain the final score.
We report per-category scores and their macro average, using
\texttt{Qwen2.5-32B-Instruct}\footnote{\url{https://huggingface.co/Qwen/Qwen2.5-32B-Instruct}} at temperature $0.0$ as the judge to stay consistent with prior work~\citep{lampqa,iap}.
More details are provided in Appendix~\ref{apd_dataset}.

\par
\textbf{Baselines.}
We compare \ours with the following baselines, with more details in Appendix~\ref{apd_baseline}:
\begin{itemize}[leftmargin=*, topsep=0pt, itemsep=0pt]
    \item \textbf{Non-Perso}: A non-personalized baseline which answers the question without any user-specific information in the model input.
    \item \textbf{RAG}~\citep{lamp}: A method that retrieves the most relevant entries from the user profile and uses them as instructional context for personalization.
    \item \textbf{SFT}~\citep{sft}: This method samples five candidate responses per question, retains the one with the highest personalized score as a silver label, and fine-tunes on the resulting pairs, thereby using the aspects only as a selection criterion.
    \item \textbf{PlanPers}~\citep{lampqa}: This method infers the aspects a user is likely to care about before answering using a planner fine-tuned on the training-split aspects.
    \item \textbf{IAP}~\citep{iap}: This method extends DAPO~\citep{dapo} by steering the model to elicit the latent intent before answering and reward responses based on the rubric aspects.
\end{itemize}

\par
\textbf{Backbone Models.}
We evaluate \ours on four representative backbone models: \texttt{Gemma2-9B}\footnote{\url{https://huggingface.co/google/gemma-2-9b-it}}, \texttt{Qwen2.5-7B}\footnote{\url{https://huggingface.co/Qwen/Qwen2.5-7B-Instruct}}, \texttt{Qwen2.5-14B}\footnote{\url{https://huggingface.co/Qwen/Qwen2.5-14B-Instruct}}, and \texttt{Qwen3-4B}\footnote{\url{https://huggingface.co/Qwen/Qwen3-4B-Instruct-2507}}.
For every method, the same backbone is used as the underlying model to ensure a fair comparison, and in \ours\ each backbone serves as both the teacher and the student.

\par
\textbf{Implementation Details.}
Following the official LaMP-QA~\citep{lampqa} protocol, both the student and teacher receive the top-$10$ user-profile entries ranked by Contriever~\citep{contriever}, while only the teacher additionally observes the target rubric aspects.
For each backbone, the student and frozen teacher share the same initialization, while the student is optimized through full-parameter fine-tuning with the full-vocabulary forward KL objective.
RTV is conducted separately for each backbone using its corresponding rubric-informed teacher.
We train \ours for 1 epoch and select the checkpoint with the highest score on the validation set for testing.
More details are provided in Appendix~\ref{apd_impl}.

\subsection{Main Results}

\setlength{\aboverulesep}{0pt}
\setlength{\belowrulesep}{0pt}
\setlength{\extrarowheight}{.75ex}
\begin{table*}[t]
\centering
\caption{
Performance comparison between the baselines and \ours across four backbones of varying scale on LaMP-QA.
The best results are highlighted in \textbf{bold}, and the second-best results are \underline{underlined}.
Higher values indicate better performance.
}
\vspace{-2mm}
\setlength{\tabcolsep}{6pt}
\renewcommand{\arraystretch}{1.04}
\resizebox{\textwidth}{!}{%
\begin{tabular}{ll cccccc}
\toprule
\textbf{Backbone} & \textbf{Category}
& Non-Perso & RAG & SFT & PlanPers & IAP & \textbf{\ours} \\
\midrule
\multirow{4}{*}{\textbf{\texttt{Gemma2-9B}}}
& A\&E  & 0.2206 & 0.3046 & 0.3178 & \underline{0.3589} & 0.2758 & \cellcolor{oursgray}\textbf{0.4372} \\
& L\&PD & 0.4030 & 0.4260 & 0.4447 & \underline{0.4792} & 0.4417 & \cellcolor{oursgray}\textbf{0.5606} \\
& S\&C  & 0.4253 & 0.4846 & 0.4976 & \underline{0.5456} & 0.4750 & \cellcolor{oursgray}\textbf{0.5787} \\
\cdashline{2-8}
& \textbf{Avg. ($\uparrow$)} & 0.3496 & 0.4051 & 0.4200 & \underline{0.4612} & 0.3975 & \cellcolor{oursgray}\textbf{0.5255} \\
\midrule
\multirow{4}{*}{\textbf{\texttt{Qwen2.5-7B}}}
& A\&E  & 0.3285 & 0.3363 & 0.3602 & 0.3748 & \underline{0.3859} & \cellcolor{oursgray}\textbf{0.4408} \\
& L\&PD & 0.4651 & 0.4440 & 0.4625 & 0.4812 & \underline{0.5207} & \cellcolor{oursgray}\textbf{0.5708} \\
& S\&C  & 0.4723 & 0.5027 & 0.5089 & 0.5359 & \underline{0.5457} & \cellcolor{oursgray}\textbf{0.5955} \\
\cdashline{2-8}
& \textbf{Avg. ($\uparrow$)} & 0.4220 & 0.4277 & 0.4439 & 0.4640 & \underline{0.4841} & \cellcolor{oursgray}\textbf{0.5357} \\
\midrule
\multirow{4}{*}{\textbf{\texttt{Qwen2.5-14B}}}
& A\&E  & 0.3379 & 0.3734 & 0.3782 & \underline{0.4384} & 0.4097 & \cellcolor{oursgray}\textbf{0.4736} \\
& L\&PD & 0.4870 & 0.4692 & 0.4782 & \underline{0.5425} & 0.5255 & \cellcolor{oursgray}\textbf{0.5888} \\
& S\&C  & 0.5030 & 0.5075 & 0.5214 & \underline{0.5965} & 0.5615 & \cellcolor{oursgray}\textbf{0.6347} \\
\cdashline{2-8}
& \textbf{Avg. ($\uparrow$)} & 0.4426 & 0.4500 & 0.4593 & \underline{0.5258} & 0.4989 & \cellcolor{oursgray}\textbf{0.5657} \\
\midrule
\multirow{4}{*}{\textbf{\texttt{Qwen3-4B}}}
& A\&E  & 0.3245 & 0.4192 & 0.4249 & \underline{0.4972} & 0.3985 & \cellcolor{oursgray}\textbf{0.5215} \\
& L\&PD & 0.5872 & 0.5346 & 0.5408 & \underline{0.6460} & 0.5698 & \cellcolor{oursgray}\textbf{0.6461} \\
& S\&C  & 0.5900 & 0.6003 & 0.6033 & \textbf{0.6978} & 0.5326 & \cellcolor{oursgray}\underline{0.6863} \\
\cdashline{2-8}
& \textbf{Avg. ($\uparrow$)} & 0.5006 & 0.5180 & 0.5230 & \underline{0.6137} & 0.5003 & \cellcolor{oursgray}\textbf{0.6180} \\
\bottomrule
\end{tabular}}
\label{tab_main_results}
\vspace{-2mm}
\end{table*}

We begin by evaluating the overall performance of all compared methods on LaMP-QA. 
Experimental results are presented in Table~\ref{tab_main_results}, from which we draw the following observations:
\begin{itemize}[leftmargin=*, topsep=0pt, itemsep=0pt]
    \item \textbf{User profiles provide valuable signals for personalized question answering.}
    Across the four backbones, incorporating user profiles generally improves performance over the Non-Perso setting.
    This demonstrates that historical user information provides useful signals for understanding user-specific preferences and information needs.
    \item \textbf{Directly enhancing model capability yields suboptimal personalization gains.}
    SFT yields consistent but modest gains over RAG, as its silver labels are generated by the backbone itself and are therefore constrained by its existing personalization capability.
    This limits the additional user-specific guidance that SFT can provide.
    \item \textbf{Rubric aspects offer stronger supervision, yet coarse-grained utilization limits their potential.}
    PlanPers generally achieves substantial gains, while IAP performs strongly on certain backbones but remains unstable across others.
    Both methods leverage rubric aspects only at a coarse granularity, either as supervision for planner training or as scalar rewards, highlighting the need for finer-grained supervision to better exploit rubric information.
    \item \textbf{\ours consistently achieves state-of-the-art performance.}
    \ours consistently obtains the highest average score on all four backbones with diverse model families and scales.
    For instance, on the \texttt{Gemma2-9B} backbone, \ours achieves an average relative improvement of $14$\% over the strongest baseline.
    These results highlight the effectiveness of our rubric-aware on-policy self-distillation framework and underscore the importance of leveraging user-specific rubric aspects as fine-grained, token-level supervision.
\end{itemize}

\subsection{Framework Analyses}

In this section, we conduct additional experiments to further study the design and effectiveness of our proposed \ours.

\begin{wrapfigure}{r}{0.6\linewidth}
    \vspace{-1mm}
    \centering
    \includegraphics[width=\linewidth]{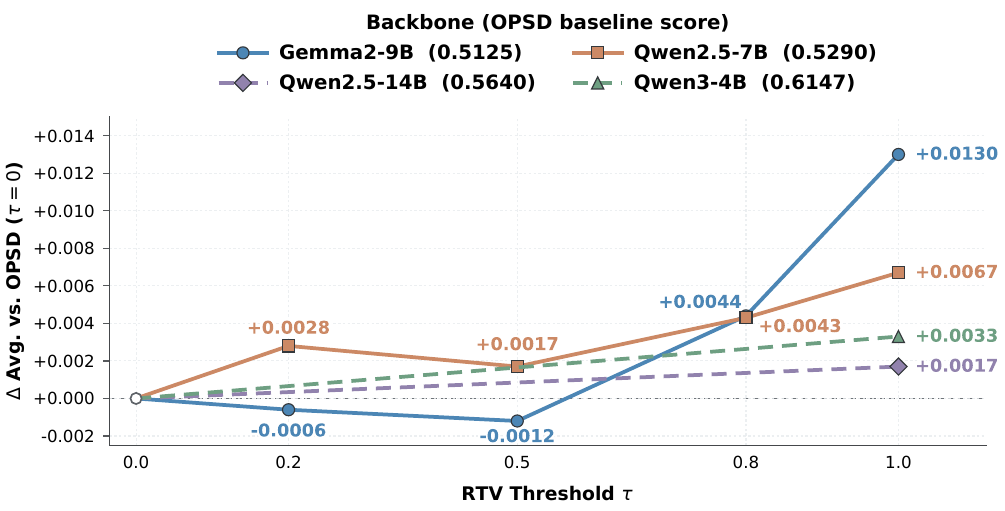}
    \vspace{-4mm}
    \caption{Effect of RTV and its threshold $\tau$.}
    \label{fig_rtv_threshold}
    \vspace{-3mm}
\end{wrapfigure}

\par
\textbf{RTV Ablation and Threshold Analysis.}
We study the effect of our introduced RTV and its threshold $\tau$.
Here, $\tau=0$ denotes vanilla OPSD without teacher validation, while larger values impose stricter requirements on teacher reliability.
We evaluate $\tau\in\{0,0.2,0.5,0.8,1.0\}$ on \texttt{Gemma2-9B} and \texttt{Qwen2.5-7B}, and compare $\tau=0$ and $\tau=1$ on \texttt{Qwen2.5-14B} and \texttt{Qwen3-4B}, with all other settings unchanged.

As illustrated in Figure~\ref{fig_rtv_threshold}, RTV with $\tau=1$ consistently improves performance across all four backbones.
For \texttt{Gemma2-9B} and \texttt{Qwen2.5-7B}, lower thresholds yield limited and non-monotonic gains, while stricter validation generally leads to better performance.
This suggests that lenient validation may retain unreliable teacher signals with incomplete rubric coverage, whereas requiring full coverage provides more reliable supervision for distillation.
Overall, these results demonstrate the effectiveness of RTV and support $\tau=1$ as our default setting.
We provide a more detailed analysis of RTV in Appendix~\ref{apd_rtv}.

\begin{wrapfigure}{r}{0.71\linewidth}
    \vspace{-1mm}
    \centering
    \includegraphics[width=\linewidth]{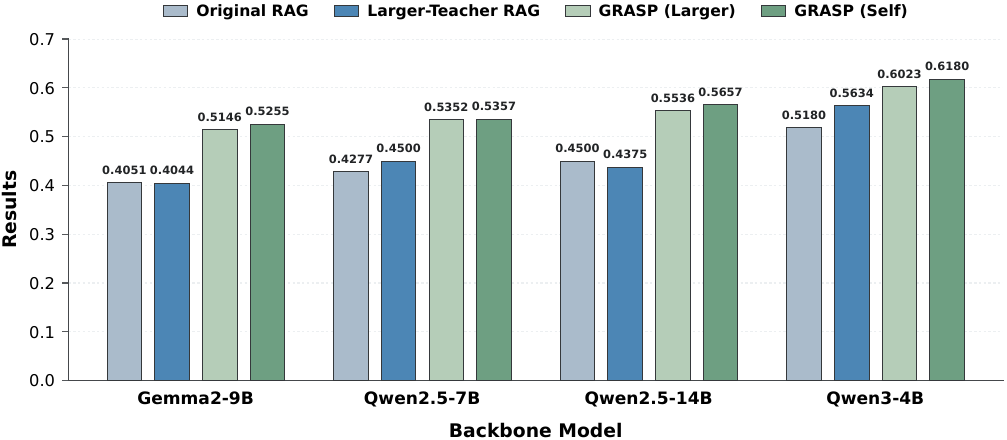}
    \vspace{-4mm}
    \caption{Teacher capability and distillation effectiveness.}
    \label{fig_teacher_capability}
    \vspace{-2mm}
\end{wrapfigure}

\par
\textbf{Teacher Capability and Distillation Effectiveness.}
To investigate whether larger models can further improve personalization, we compare each backbone with its larger counterpart specified in Appendix~\ref{apd_impl} under both standard RAG and our self-distillation framework.
Under RAG, we directly compare the two model sizes to assess the benefit of increased model capacity.
For \ours, we keep the backbone as the student and use either itself or its larger counterpart as the rubric-informed teacher.
We report the macro-average score across the three categories.

As shown in Figure~\ref{fig_teacher_capability}, simply scaling up the model under RAG yields inconsistent gains and can even degrade performance.
Similarly, using a larger teacher in \ours provides no clear advantage over self-distillation.
These results suggest that increased model or teacher capacity alone is not the key to effective PQA.
Instead, the gains primarily come from leveraging user-specific rubric aspects and effectively transferring their guidance to the student, further supporting the core design of \ours.

\par
\textbf{Effect of User-Specific Rubric Alignment.}
To examine whether the effectiveness of \ours arises from the question-specific information encoded in the user-specific rubric aspects, rather than simply from additional teacher context, we construct a shuffled-rubric variant of \ours.
Specifically, we keep the question and user profile unchanged while randomly shuffling rubric sets across training instances.
This preserves the overall distribution of rubric aspects but breaks their instance-level alignment with the corresponding question and user profile.
All other settings remain unchanged.

As shown in Table~\ref{tab_rubric_alignment}, shuffled rubrics lead to a substantial performance drop.
This result demonstrates that the benefit of rubric-aware self-distillation stems not from exposing the teacher to arbitrary additional context, but from the question-specific semantic targets encoded in the rubric aspects that guide the student toward the current user’s information needs.
Misaligned rubrics steer the teacher toward irrelevant continuations, making the transferred token-level preferences ineffective for personalization and highlighting the importance of accurate user–question–rubric alignment.

\begin{table}[t]
\centering
\small
\setlength{\tabcolsep}{4pt}
\renewcommand{\arraystretch}{0.9}
\caption{
Effect of user-specific rubric alignment under \ours.
\textbf{Shuffled} uses rubrics from other questions, while \textbf{Matched} uses the corresponding rubrics.
}
\vspace{-3mm}
\label{tab_rubric_alignment}

\resizebox{0.85\linewidth}{!}{%
\begin{tabular}{lcccccccc}
\toprule
\multirow{2}{*}{\textbf{Category}}
& \multicolumn{2}{c}{\textbf{Gemma2-9B}}
& \multicolumn{2}{c}{\textbf{Qwen2.5-7B}}
& \multicolumn{2}{c}{\textbf{Qwen2.5-14B}}
& \multicolumn{2}{c}{\textbf{Qwen3-4B}} \\
\cmidrule(lr){2-3}
\cmidrule(lr){4-5}
\cmidrule(lr){6-7}
\cmidrule(lr){8-9}
& \textbf{Shuffled} & \textbf{Matched}
& \textbf{Shuffled} & \textbf{Matched}
& \textbf{Shuffled} & \textbf{Matched}
& \textbf{Shuffled} & \textbf{Matched} \\
\midrule
A\&E
& 0.3016 & \textbf{0.4372}
& 0.3612 & \textbf{0.4408}
& 0.3477 & \textbf{0.4736}
& 0.4246 & \textbf{0.5215} \\

L\&PD
& 0.4385 & \textbf{0.5606}
& 0.4488 & \textbf{0.5708}
& 0.4505 & \textbf{0.5888}
& 0.5290 & \textbf{0.6461} \\

S\&C
& 0.4745 & \textbf{0.5787}
& 0.4643 & \textbf{0.5955}
& 0.4744 & \textbf{0.6347}
& 0.5953 & \textbf{0.6863} \\
\midrule
\rowcolor{oursgray}
Avg.
& 0.4049 & \textbf{0.5255}
& 0.4248 & \textbf{0.5357}
& 0.4242 & \textbf{0.5657}
& 0.5163 & \textbf{0.6180} \\
\bottomrule
\end{tabular}%
}

\vspace{-2mm}
\end{table}

\textbf{Rubric-Induced Token Distribution Shifts.}
We next examine whether the gains of \ours arise from its intended mechanism: transferring the token-level effect of privileged rubric information into a rubric-free student.
We conduct this analysis on \texttt{Qwen2.5-7B} using $90$ held-out responses.
Specifically, for each question and user profile, we first sample a response from the student before training. 
At every answer position, we then feed the same sampled prefix to three models: the student before training, the student after training, and the rubric-conditioned teacher. 
This matched-prefix setup ensures that the observed differences arise from changes in the next-token distribution rather than from the models generating different preceding text.

\begin{figure*}[!t]
    \centering
    \includegraphics[width=1.0\linewidth]{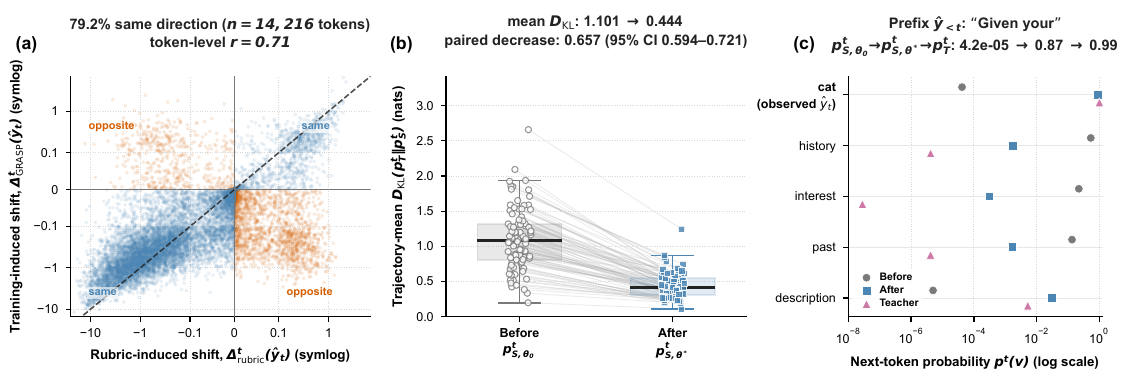}
    \vspace{-4mm}
    \caption{
GRASP internalizes rubric-induced token-distribution shifts.
\textbf{(a)} Token-level alignment between rubric-induced and training-induced probability shifts.
Blue and orange denote matching and opposing directions; the dashed line denotes equal shifts.
\textbf{(b)} Response-level forward KL from the rubric-conditioned teacher to the student before and after training.
\textbf{(c)} Representative next-token distributions before training, after training, and with rubric conditioning.
    }
    \label{fig_rubric_shift}
    \vspace{-4mm}
\end{figure*}

Figure~\ref{fig_rubric_shift}(a) compares rubric-induced and training-induced probability shifts for each observed token. 
The two shifts agree in direction for $79.2\%$ of 14,216 tokens, with a correlation of $r=0.71$, indicating that \ours learns to promote rubric-favored tokens and suppress rubric-disfavored ones.
Figure~\ref{fig_rubric_shift}(b) extends this analysis to the full vocabulary. 
The response-level forward KL between teacher and student decreases from $1.101$ to $0.444$ on average after training, with all 90 responses moving closer to the rubric-conditioned teacher.
Figure~\ref{fig_rubric_shift}(c) shows a representative example:
after the prefix ``Given your,'' training raises the probability of the observed token ``cat'' from $4.2\times10^{-5}$ to $0.87$, close to the teacher probability of $0.99$, while generic alternatives are suppressed.
Together, these results suggest that \ours transfers the rubric's steering effect into the student itself: training reshapes its token distributions toward those induced by rubric conditioning, enabling rubric-guided generation even when the rubric is absent at inference time.

\section{Conclusion}

In this work, we introduced \ours, a rubric-aware on-policy self-distillation framework for personalized question answering.
\ours transforms user-specific rubric aspects into fine-grained, token-level supervision by aligning a rubric-free student with a frozen rubric-informed teacher along student-generated trajectories.
We further proposed Rubric-based Teacher Validation to improve the reliability of teacher guidance.
Experiments on LaMP-QA across four diverse backbones show that \ours consistently achieves state-of-the-art performance, while further analyses confirm the importance of user-specific rubric aspects and reliable supervision.
Overall, our findings demonstrate the value of fine-grained distributional supervision for personalization.
To the best of our knowledge, this work is the first to introduce on-policy self-distillation as a post-training approach to personalization, offering a new paradigm for adapting language models to individual users.

\bibliography{iclr2027_conference}
\bibliographystyle{iclr2027_conference}

\newpage
\appendix

\section{Dataset Details}\label{apd_dataset}

We conduct all experiments on LaMP-QA\footnote{\url{https://huggingface.co/datasets/alireza7/LaMP-QA}}~\citep{lampqa}, a benchmark for personalized long-form question answering.
The benchmark is constructed by treating a user's current question as the input, their previously asked questions as the user profile, and the key aspects extracted from the question narrative as the evaluation rubric.
Questions are grouped into three categories: \emph{Arts \& Entertainment}, \emph{Lifestyle \& Personal Development}, and \emph{Society \& Culture}.
Detailed statistics are reported in Table~\ref{tab_dataset_stats}.

\begin{table*}[t]
\centering
\caption{Data statistics of the three LaMP-QA categories.}
\vspace{-2mm}
\label{tab_dataset_stats}
\setlength{\tabcolsep}{2pt}
\renewcommand{\arraystretch}{1.4}
\resizebox{1\textwidth}{!}{%
\begin{tabular}{lccccccccc}
\toprule
\multirow{2}{*}{\textbf{Statistic}}
& \multicolumn{3}{c}{\textbf{Arts \& Entertainment}}
& \multicolumn{3}{c}{\textbf{Lifestyle \& Personal Development}}
& \multicolumn{3}{c}{\textbf{Society \& Culture}} \\
\cmidrule(lr){2-4} \cmidrule(lr){5-7} \cmidrule(lr){8-10}
& Train & Valid & Test & Train & Valid & Test & Train & Valid & Test \\
\midrule
\#Questions (users)
& 9{,}349 & 801 & 767
& 7{,}370 & 892 & 989
& 7{,}614 & 810 & 1{,}074 \\

\#Rubric aspects
& 2.7{\scriptsize$\pm$0.9} & 4.7{\scriptsize$\pm$1.2} & 4.6{\scriptsize$\pm$1.2}
& 3.1{\scriptsize$\pm$1.0} & 5.1{\scriptsize$\pm$1.1} & 5.1{\scriptsize$\pm$1.2}
& 2.9{\scriptsize$\pm$0.9} & 4.8{\scriptsize$\pm$1.1} & 4.8{\scriptsize$\pm$1.0} \\

Profile size
& 106.7{\scriptsize$\pm$127.3} & 129.0{\scriptsize$\pm$183.7} & 159.1{\scriptsize$\pm$203.0}
& 116.6{\scriptsize$\pm$162.0} & 98.2{\scriptsize$\pm$198.6} & 111.6{\scriptsize$\pm$220.3}
& 141.3{\scriptsize$\pm$194.7} & 110.5{\scriptsize$\pm$210.6} & 115.8{\scriptsize$\pm$203.6} \\

Question length
& 13.0{\scriptsize$\pm$2.9} & 10.6{\scriptsize$\pm$4.0} & 10.0{\scriptsize$\pm$3.8}
& 13.6{\scriptsize$\pm$3.3} & 11.3{\scriptsize$\pm$4.4} & 11.6{\scriptsize$\pm$4.6}
& 14.2{\scriptsize$\pm$3.6} & 12.1{\scriptsize$\pm$4.9} & 12.9{\scriptsize$\pm$5.4} \\
\bottomrule
\end{tabular}}
\vspace{-2mm}
\end{table*}

\section{Baseline Details}\label{apd_baseline}

In this section, we further introduce each baseline method in detail. The comparison between different baselines and our method is shown in Table~\ref{method_comparison}.

\begin{itemize}[leftmargin=*, topsep=0pt, itemsep=0pt]
    \item \textbf{Non-Perso}:
This baseline directly applies the unmodified backbone to the current question without providing any user-profile information.
It therefore measures the performance of general question answering in the absence of personalization.
    \item \textbf{RAG}:
This baseline augments the current question with the top-$10$ entries retrieved from the user profile and feeds them to the unmodified backbone for response generation.
It measures the personalization gains achieved by directly incorporating relevant user history as additional context.
    \item \textbf{SFT}:
We construct silver-labeled training data through best-of-five rejection sampling.
For each training instance, the corresponding backbone generates five candidate responses conditioned on the question and retrieved user profile.
We evaluate each candidate against the target rubric aspects, following the same per-aspect scoring procedure used for test-set evaluation, and retain the candidate with the highest average score.
The backbone is then fine-tuned on the resulting question--profile--response pairs.
    \item \textbf{PlanPers}:
We follow the two-stage PlanPers pipeline.
First, a planner initialized from \texttt{Qwen2.5-7B-Instruct} is fine-tuned with LoRA to predict the rubric aspect titles from the question and retrieved user profile.
At inference time, the planner first predicts a set of relevant aspects for each question.
The corresponding frozen backbone then generates the final response conditioned on the question, retrieved user profile, and predicted aspects.
    \item \textbf{IAP}:
We implement IAP by prompting the policy to first articulate the inferred user intent and then generate the final personalized response.
The policy is optimized with DAPO.
Following the original formulation, its reward combines rubric-based personalization quality, a contrastive term relative to an intent-free reference response, and a penalty on excessively long intent reasoning.
\end{itemize}

\begin{table*}[t]
    \caption{
    We provide a comparison between the different baseline methods and our proposed \ours, focusing on the following aspects:
    (1) whether the method conditions on the user profile, (2) whether it trains the backbone model, (3) whether it exploits user-specific rubric aspects, and (4) how rubric-related supervision is incorporated and at what granularity.
    We further rank these supervision mechanisms by granularity, with \#1 denoting the finest-grained supervision.
    }
    \vspace{-2mm}
    \label{method_comparison}
    \centering
    \renewcommand{\arraystretch}{1.0}
    \resizebox{0.9\textwidth}{!}{
    \begin{tabular}{c cc  cc  c}
        \toprule
        \textbf{Methods} & \textbf{User Profile} & \textbf{Training} & \textbf{Rubric Aspects} & \textbf{Supervision Granularity} \\
        \midrule
        Non-Perso  & \xmark & \xmark & \xmark & --- \\
        RAG        & \cmark & \xmark & \xmark & --- \\
        SFT        & \cmark & \cmark & \cmark & Data filtering~(\#4) \\
        PlanPers   & \cmark & \xmark & \cmark & Plan prediction~(\#3) \\
        IAP        & \cmark & \cmark & \cmark & RL reward~(\#2) \\
        \midrule\rowcolor{oursgray}
        \ours~(ours)        & \cmark & \cmark & \cmark & Token-level KL~(\#1) \\
        \bottomrule
    \end{tabular}
    }
    \vspace{-2mm}
\end{table*}

\section{Implementation Details}\label{apd_impl}

\par
\textbf{Rubric-based Teacher Validation.}
RTV is performed separately for each backbone using its corresponding frozen teacher.
For every training instance, the teacher generates one validation response conditioned on the question, user profile, and target rubric aspects.
These responses are generated with a temperature of $0.1$, top-$p$ of $0.95$, and a maximum length of $1024$ tokens.
To maintain consistency with the test-set evaluation protocol, we use \texttt{Qwen2.5-32B-Instruct} as the rubric judge and follow the same per-aspect scoring procedure at temperature $0.0$.
The normalized per-aspect scores are averaged to obtain the teacher reliability score $s_T$.

\par
\textbf{Model training.}
For each backbone, the student and frozen teacher share the same initialization.
The teacher remains frozen throughout training, while the student is optimized through full-parameter fine-tuning.
The maximum sequence length is $6{,}144$ tokens, including a maximum of $512$ completion tokens.
We use AdamW~\citep{adamw} with a learning rate of $5\times10^{-6}$, weight decay of $0.1$, and a cosine learning-rate schedule.
The per-device batch size is set to $1$, and we set the gradient-accumulation factor to 4 for \texttt{Qwen2.5-7B} and \texttt{Gemma2-9B} and to 8 for \texttt{Qwen2.5-14B} and \texttt{Qwen3-4B}.
We employ DeepSpeed ZeRO-3, bfloat16 precision, and gradient checkpointing for memory-efficient training, and clip the gradient norm to $1.0$.
For each training instance, we sample a response on-policy from the current student using a temperature of $0.9$, top-$p$ of $0.9$, and top-$k$ of $50$.
At each position of the sampled response, the teacher and student evaluate the same student-generated prefix.
We compute the forward KL divergence over the full vocabulary with a distillation temperature of $1.0$.

\par
\textbf{Inference.}
Following the official LaMP-QA protocol, evaluation responses are generated using the top-$10$ retrieved profile entries, with a temperature of $0.1$, top-$p$ of $0.95$, and a maximum generation length of $2048$ tokens.

\par
\textbf{Running environment.}
We implement \ours based on the ms-swift\footnote{\url{https://github.com/modelscope/ms-swift}} framework.
All experiments are conducted on 8 NVIDIA H800 GPUs.

\par
\textbf{Larger Models for Comparison.}
For the teacher-capability analysis, we pair each student backbone with a larger model from the same family:
\texttt{Gemma-2-9B-it} with \texttt{Gemma-2-27B-it},
\texttt{Qwen2.5-7B-Instruct} with \texttt{Qwen2.5-14B-Instruct},
\texttt{Qwen2.5-14B-Instruct} with \texttt{Qwen2.5-32B-Instruct}, and
\texttt{Qwen3-4B-Instruct-2507} with \texttt{Qwen3-30B-A3B-Instruct-2507}.
Each larger model is evaluated both as a standalone RAG baseline and as a rubric-informed teacher for distilling its paired student.

\section{In-Depth Analysis}

\begin{figure*}[t]
\centering
\begin{minipage}[c]{0.48\textwidth}
  \centering
  \includegraphics[width=0.96\linewidth]{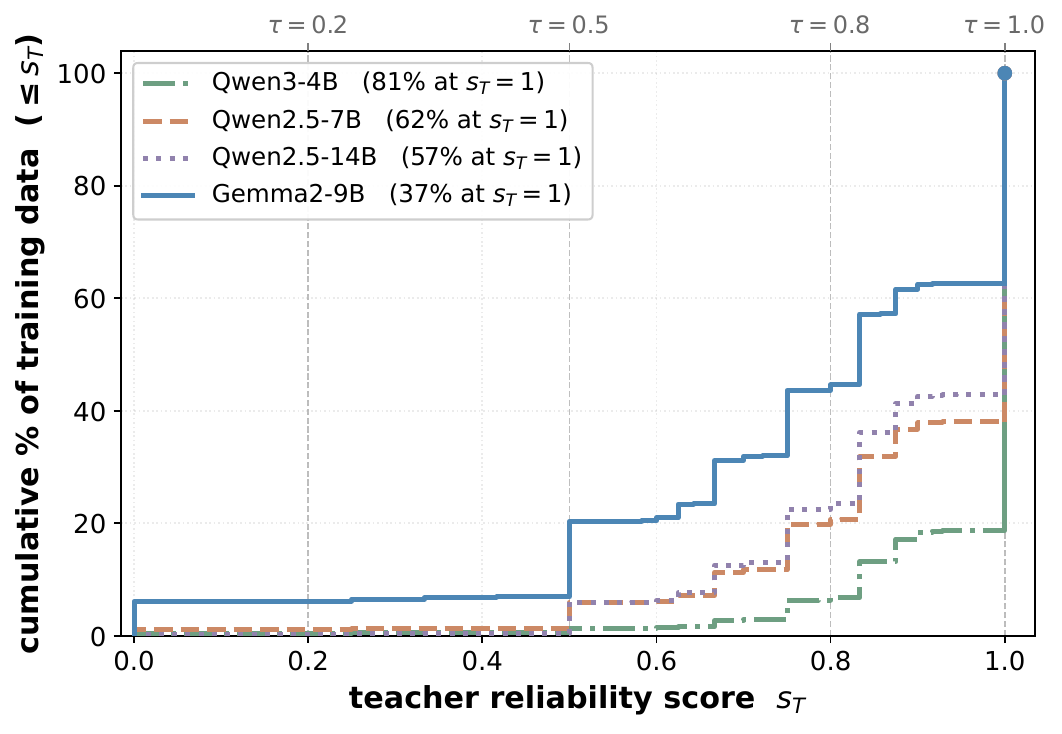}
  \vspace{-2mm}
  \caption{
Empirical distributions of teacher reliability scores across backbones. 
  }
  \label{fig_rtv_ecdf}
\end{minipage}
\hfill
\begin{minipage}[c]{0.48\textwidth}
  \centering
  \small
  \setlength{\tabcolsep}{5pt}
  \renewcommand{\arraystretch}{1.15}
  \captionof{table}{
    Training cost of \ours across backbones, including the number of training samples, training steps, and wall-clock time.
  }
  \vspace{-2mm}
  \begin{tabular}{lccc}
    \toprule
    \textbf{Backbone} & \textbf{\#Samples} & \textbf{\#Steps} & \textbf{Time} \\
    \midrule
    \texttt{Gemma2-9B}    & \phantom{0}9{,}050 & 283 & \phantom{0}6.7\,h \\
    \texttt{Qwen2.5-7B}    & 15{,}011 & 470 & \phantom{0}6.7\,h \\
    \texttt{Qwen2.5-14B}   & 13{,}847 & 217 & 10.9\,h \\
    \texttt{Qwen3-4B} & 19{,}700 & 308 & 11.9\,h \\
    \bottomrule
  \end{tabular}
  \label{tab_train_efficiency}
\end{minipage}
\end{figure*}

\subsection{RTV Ablation and Threshold Analysis}\label{apd_rtv}

To understand why \texttt{Gemma2-9B} benefits more substantially from increasing the RTV threshold than the other backbones, we examine the distribution of teacher reliability scores across the training set.
Because RTV retains only instances satisfying $s_T \geq \tau$, the cumulative proportion below each threshold indicates how much potentially unreliable supervision is filtered out.

As shown in Figure~\ref{fig_rtv_ecdf}, \texttt{Gemma2-9B} exhibits a considerably heavier tail in the low- and intermediate-reliability regions. 
Only $37\%$ of its training instances attain $s_T=1$, compared with $57\%$, $62\%$, and $81\%$ for \texttt{Qwen2.5-14B}, \texttt{Qwen2.5-7B}, and \texttt{Qwen3-4B}, respectively.
Consequently, at a permissive threshold, many \texttt{Gemma2-9B} instances whose teacher responses only partially cover the target rubric aspects remain in the training set.
Raising $\tau$ filters out a substantially larger proportion of such unreliable supervision, providing a distributional explanation for the more pronounced performance gains of \texttt{Gemma2-9B} at higher thresholds in Figure~\ref{fig_rtv_threshold}.
By contrast, the reliability distributions of the other teachers are already more concentrated at $s_T=1$; increasing $\tau$ therefore changes their retained training sets less substantially and leads to smaller incremental gains.
Overall, these results suggest that the effectiveness of RTV is closely related to the reliability distribution of the rubric-informed teacher, with stricter filtering being particularly beneficial when the distribution contains substantial low-score mass.

\subsection{Training Efficiency}

Although \ours involves on-policy response sampling and full-vocabulary distillation from a frozen teacher, its training cost remains practical. 
Table~\ref{tab_train_efficiency} reports the number of samples retained after RTV, the corresponding optimization steps, and the wall-clock training time.

Across all four backbones, training requires only $217$--$470$ steps and finishes within $6.7$--$11.9$ hours.
At inference time, \ours takes the same input as RAG, consisting only of the question and retrieved user profile, without requiring rubric aspects or the teacher model.
It therefore introduces no additional inference cost over standard RAG.
Overall, \ours provides fine-grained supervision with manageable training costs and no additional inference overhead.

\begin{table}[t]
\centering
\small
\setlength{\tabcolsep}{4pt}
\renewcommand{\arraystretch}{0.9}
\caption{
Effect of rubric completeness as privileged teacher information.
\textbf{Single} provides one randomly sampled target aspect, whereas \textbf{Full} provides the complete set of target aspects.
}
\vspace{-2mm}
\label{tab_rubric_completeness}

\resizebox{0.88\linewidth}{!}{%
\begin{tabular}{lcccccccc}
\toprule
\multirow{2}{*}{\textbf{Category}}
& \multicolumn{2}{c}{\textbf{Gemma2-9B}}
& \multicolumn{2}{c}{\textbf{Qwen2.5-7B}}
& \multicolumn{2}{c}{\textbf{Qwen2.5-14B}}
& \multicolumn{2}{c}{\textbf{Qwen3-4B}} \\
\cmidrule(lr){2-3}
\cmidrule(lr){4-5}
\cmidrule(lr){6-7}
\cmidrule(lr){8-9}
& \textbf{Single} & \textbf{Full}
& \textbf{Single} & \textbf{Full}
& \textbf{Single} & \textbf{Full}
& \textbf{Single} & \textbf{Full} \\
\midrule
A\&E
& 0.4117 & \textbf{0.4372}
& 0.4196 & \textbf{0.4408}
& 0.4448 & \textbf{0.4736}
& 0.4980 & \textbf{0.5215} \\

L\&PD
& 0.5191 & \textbf{0.5606}
& 0.5401 & \textbf{0.5708}
& 0.5420 & \textbf{0.5888}
& 0.6159 & \textbf{0.6461} \\

S\&C
& 0.5559 & \textbf{0.5787}
& 0.5604 & \textbf{0.5955}
& 0.5853 & \textbf{0.6347}
& 0.6651 & \textbf{0.6863} \\
\midrule
\rowcolor{oursgray}
Avg.
& 0.4956 & \textbf{0.5255}
& 0.5067 & \textbf{0.5357}
& 0.5240 & \textbf{0.5657}
& 0.5930 & \textbf{0.6180} \\
\bottomrule
\end{tabular}%
}

\vspace{-2mm}
\end{table}

\begin{figure*}[!t]
    \centering
    \includegraphics[width=1\linewidth]{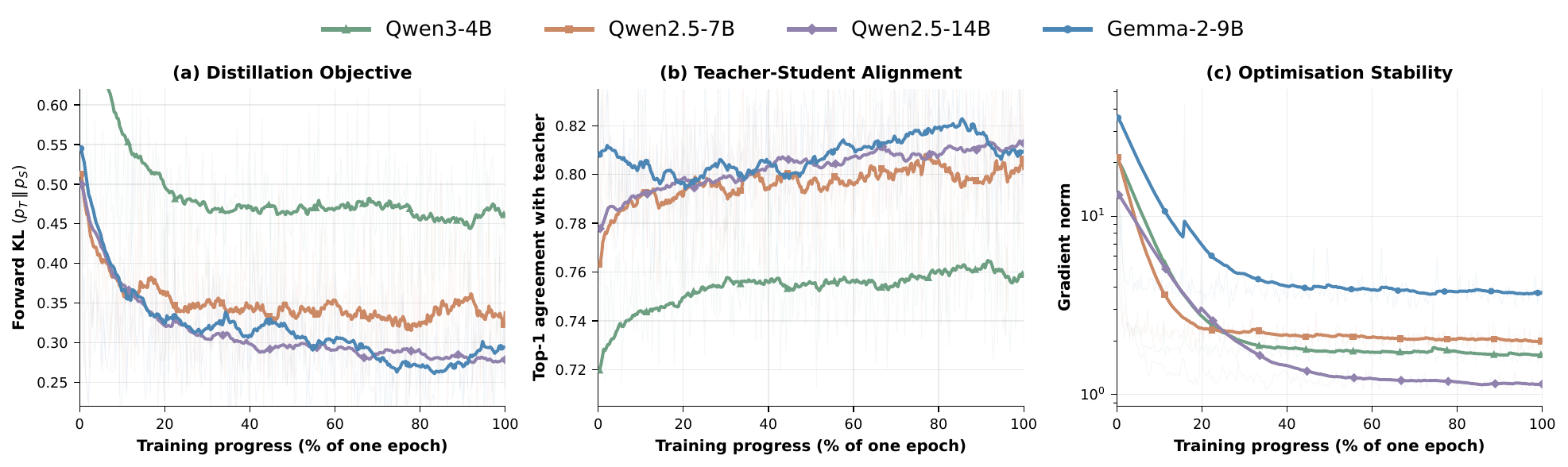}
    \vspace{-6mm}
\caption{
Training dynamics of \ours across the four backbones.
(a) Forward KL $(p_T \,\|\, p_S)$, the objective OPSD minimizes.
(b) Fraction of positions where the student and teacher share the same top-$1$ token.
(c) Gradient norm before clipping, on a log scale.
Faint traces are raw
per-step values; solid lines are an exponential moving average.
}
    \label{training_dynamics}
    \vspace{-3mm}
\end{figure*}

\subsection{Effect of Rubric Completeness}

To investigate how the completeness of privileged rubric information affects distillation, we construct a single-aspect variant of \ours.
For each training instance, the \textbf{Single} setting provides the rubric-informed teacher with only one aspect randomly sampled from $E_{x_u}$, whereas the \textbf{Full} setting provides the complete set of target aspects.
All other training configurations remain unchanged.
This comparison allows us to determine whether an individual aspect is sufficient for effective supervision or whether the teacher benefits from jointly observing the user's complete information needs.

As shown in Table~\ref{tab_rubric_completeness}, using the complete rubric set consistently outperforms the single-aspect setting across all categories and backbones.
Providing only one aspect gives the teacher a partial view of the user's information needs, resulting in incomplete supervision.
In contrast, the full rubric set captures complementary information across aspects and enables more comprehensive personalized guidance.
These results demonstrate that the effectiveness of \ours depends not only on the availability of rubric information but also on its completeness.

\subsection{Training Dynamics}

Figure~\ref{training_dynamics} tracks three training-time signals over the single epoch.
Together, they reveal rapid convergence: the forward KL decreases sharply, the teacher--student top-1 agreement increases, and the gradient norm drops by approximately one order of magnitude.
Most of these changes occur within the first $20$-$30\%$ of training, after which all three curves gradually stabilize.
These trends indicate stable optimization toward the rubric-informed teacher distributions across different backbones.

\section{Case Study}

To qualitatively examine whether \ours internalizes user-specific rubric information, we compare its response with that of the RAG baseline using \texttt{Qwen2.5-7B}.
As shown in Figure~\ref{fig_case_study}, both systems receive exactly the same inference-time input, consisting of the question and retrieved user profile. 
The displayed user narrative and rubric aspects are ground-truth evaluation criteria and are not provided to either system during inference.

The RAG baseline fails to cover all the user-specific requirements, whereas \ours successfully addresses all three rubric aspects, improving the aspect-coverage score from $0.33$ to $1.00$.
This example demonstrates that \ours can internalize rubric information during training and generate more comprehensively personalized responses.

\section{Prompts}\label{apd_prompts}

For reproducibility, we provide the prompt templates used for the student, the rubric-informed teacher, and the RTV judge. 
P1 is shared by the student and the RAG baseline, and contains only the current question and the top-$10$ retrieved user-profile entries. 
P2 retains the same base prompt but provides the frozen teacher with additional privileged information, including the user's elaboration and the target rubric aspects. 
This controlled construction ensures that the difference between the student and teacher distributions arises specifically from the rubric information.
P3 is used to evaluate whether a generated response covers each target aspect. Given the question, user details, generated response, and one rubric aspect, the judge assigns a coverage score from $0$ to $2$.

%
\definecolor{cshit}{HTML}{1B7F4B}      
\definecolor{csmiss}{HTML}{B3261E}     
\definecolor{cspart}{HTML}{C77700}     
\definecolor{cspi}{HTML}{FFF3CD}       
\definecolor{csrule}{HTML}{555555}

\newcommand{\vmiss}{\textcolor{csmiss}{\ding{55}}}
\newcommand{\vpart}{\textcolor{cspart}{\ding{119}}}
\newcommand{\vhit}{\textcolor{cshit}{\ding{51}}}

\begin{figure*}[t]
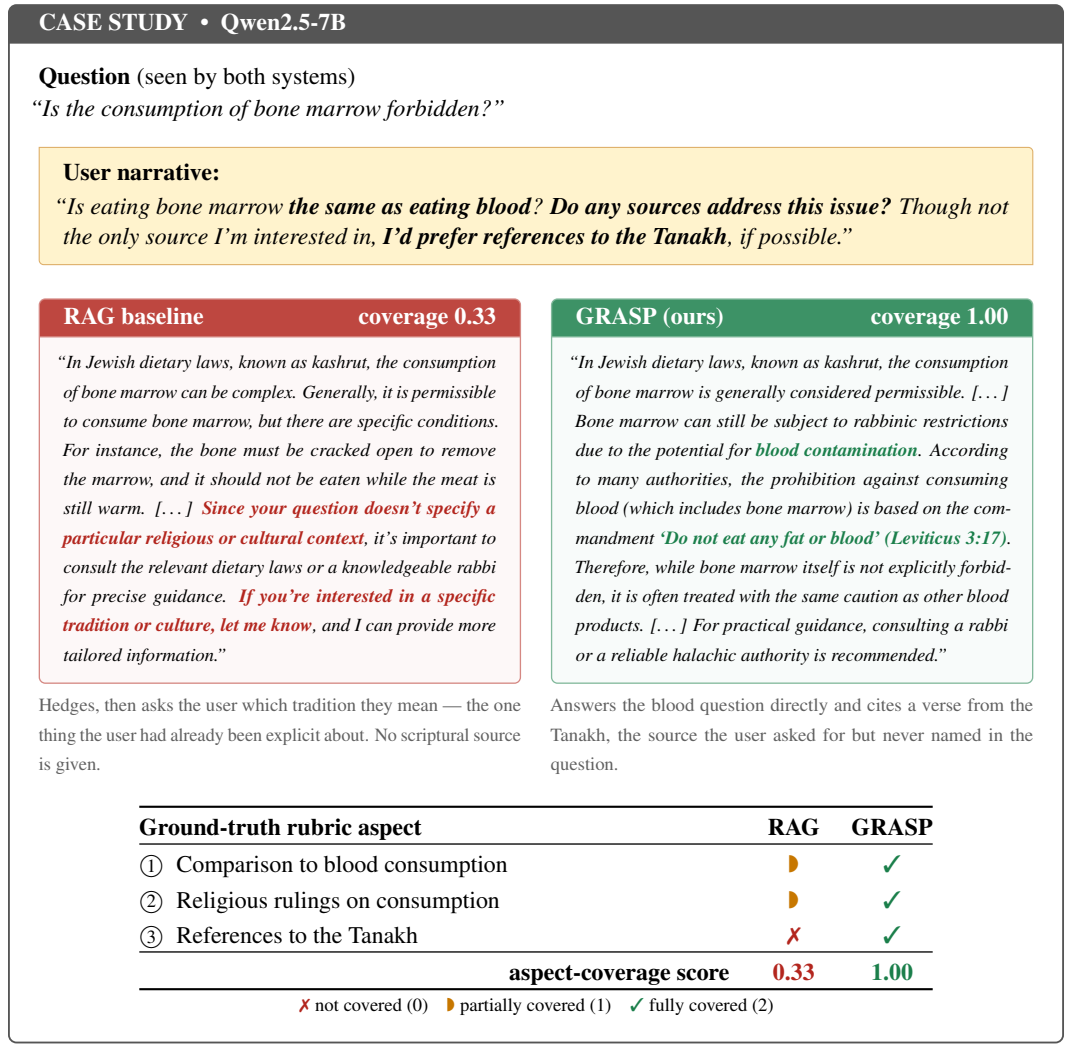

\centering
\begin{tcolorbox}[enhanced, unbreakable, colback=white, colframe=csrule,
  boxrule=0.7pt, arc=2.5pt, left=8pt, right=8pt, top=6pt, bottom=7pt,
  fonttitle=\bfseries\small, coltitle=white, colbacktitle=csrule,
  title={CASE STUDY\ \ \textbullet\ \ Qwen2.5-7B\ }]

{\small\textbf{Question}\;(seen by both systems)}\\[1pt]
{\small\itshape ``Is the consumption of bone marrow forbidden?''}

\smallskip
\begin{tcolorbox}[colback=cspi, colframe=cspart!60, boxrule=0.4pt, arc=1.5pt,
  left=6pt, right=6pt, top=3pt, bottom=3pt, sharp corners=downhill]
{\footnotesize\textbf{User narrative:}}\\[2pt]
{\footnotesize\itshape ``Is eating bone marrow \textbf{the same as eating
blood}? \textbf{Do any sources address this issue?} Though not the only source
I'm interested in, \textbf{I'd prefer references to the Tanakh}, if
possible.''}
\end{tcolorbox}

\medskip
\begin{minipage}[t]{0.485\linewidth}
\begin{tcolorbox}[colback=csmiss!3, colframe=csmiss!55, boxrule=0.5pt,
  arc=2pt, left=6pt, right=6pt, top=4pt, bottom=4pt,
  fonttitle=\bfseries\footnotesize, coltitle=white, colbacktitle=csmiss!85,
  title={RAG baseline \hfill coverage 0.33}]
{\scriptsize\itshape
``In Jewish dietary laws, known as kashrut, the consumption of bone marrow can
be complex. Generally, it is permissible to consume bone marrow, but there are
specific conditions. For instance, the bone must be cracked open to remove the
marrow, and it should not be eaten while the meat is still warm. [\ldots]
\textcolor{csmiss}{\textbf{Since your question doesn't specify a particular
religious or cultural context}}, it's important to consult the relevant dietary
laws or a knowledgeable rabbi for precise guidance.
\textcolor{csmiss}{\textbf{If you're interested in a specific tradition or
culture, let me know}}, and I can provide more tailored information.''}
\end{tcolorbox}
\vspace{-1mm}
{\scriptsize\textcolor{black!62}{Hedges, then asks the user which tradition
they mean --- the one thing the user had already been explicit about. No
scriptural source is given.}}
\end{minipage}
\hfill
\begin{minipage}[t]{0.485\linewidth}
\begin{tcolorbox}[colback=cshit!3, colframe=cshit!55, boxrule=0.5pt,
  arc=2pt, left=6pt, right=6pt, top=4pt, bottom=4pt,
  fonttitle=\bfseries\footnotesize, coltitle=white, colbacktitle=cshit!85,
  title={\ours (ours) \hfill coverage 1.00}]
{\scriptsize\itshape
``In Jewish dietary laws, known as kashrut, the consumption of bone marrow is
generally considered permissible. [\ldots] Bone marrow can still be subject to
rabbinic restrictions due to the potential for
\textcolor{cshit}{\textbf{blood contamination}}. According to many authorities,
the prohibition against consuming blood (which includes bone marrow) is based
on the commandment \textcolor{cshit}{\textbf{`Do not eat any fat or blood'
(Leviticus 3:17)}}. Therefore, while bone marrow itself is not explicitly
forbidden, it is often treated with the same caution as other blood products.
[\ldots] For practical guidance, consulting a rabbi or a reliable halachic
authority is recommended.''}
\end{tcolorbox}
\vspace{-1mm}
{\scriptsize\textcolor{black!62}{Answers the blood question directly and cites
a verse from the Tanakh, the source the user asked for but never named in the
question.}}
\end{minipage}

\medskip
\begin{center}
{\footnotesize
\begin{tabular}{@{}p{0.60\linewidth}cc@{}}
\toprule
\textbf{Ground-truth rubric aspect} & \textbf{RAG} & \textbf{\ours} \\
\midrule
\textcircled{\scriptsize 1}\; Comparison to blood consumption
  & \vpart & \vhit \\
\textcircled{\scriptsize 2}\; Religious rulings on consumption
  & \vpart & \vhit \\
\textcircled{\scriptsize 3}\; References to the Tanakh
  & \vmiss & \vhit \\
\midrule
\multicolumn{1}{@{}r}{\textbf{aspect-coverage score}\;}
  & \textcolor{csmiss}{\textbf{0.33}} & \textcolor{cshit}{\textbf{1.00}} \\
\bottomrule
\end{tabular}}\\[2pt]
{\scriptsize\vmiss\;not covered (0)\quad\vpart\;partially covered
(1)\quad\vhit\;fully covered (2)}
\end{center}

\end{tcolorbox}
\vspace{-3mm}
\caption{
A case study on LaMP-QA.
The RAG baseline misses the user's requirements, while \ours covers all three rubric aspects.
}
\label{fig_case_study}
\end{figure*}

\clearpage

\newtcolorbox{promptbox}[1]{
  breakable, enhanced, colback=gray!3, colframe=black!55, boxrule=0.5pt,
  left=5pt, right=5pt, top=4pt, bottom=4pt,
  fonttitle=\bfseries\small, title={#1}
}

\begin{promptbox}{P1 — Student prompt}
\footnotesize
\textbf{\textcolor{black!70}{[system]}}\\
You are a helpful assistant designed to generate personalized responses to
user questions. Your task is to answer a user's question from a post in a
personalized way by considering this user's past post questions and detailed
descriptions of these questions.\\[2pt]
\texttt{\# Your input:}\\
\hspace*{1em}- The user's current question from a post.\\
\hspace*{1em}- The user's past post questions and detailed descriptions of these questions.\\
\texttt{\# Your task:} Answer the user's current question in a personalized way
by considering this user's past post questions [\ldots] to learn about the
user's preferences.\\
\texttt{\# Your output:} [\ldots] a valid json object in \texttt{```json ```}
block that contains the following fields:\\
\hspace*{1em}- \texttt{personalized\_answer}: contains the personalized answer
to the user's current question [\ldots]

\smallskip\hrule\smallskip
\textbf{\textcolor{black!70}{[user]}}\\
\texttt{\# Past post questions and detailed descriptions of these questions:}\\
\texttt{\{profile\}}\hfill{\scriptsize\itshape(top-10 Contriever-ranked entries)}\\
\texttt{\# Current post question:}\\
\texttt{\{question\}}
\end{promptbox}

\vspace{4pt}

\begin{promptbox}{P2 — Teacher prompt (= P1 \textbf{+} privileged information, shaded)}
\footnotesize
\textbf{\textcolor{black!70}{[system]}}\\
{\itshape\textcolor{black!45}{(identical to P1, then appended:)}}\\[2pt]
\colorbox{yellow!22}{\parbox{0.965\linewidth}{%
\texttt{\# Additional private context (this turn only):}\\
You are additionally given private editorial notes describing what this
specific user is actually looking for: their own elaboration of the question,
and the concrete aspects they expect the answer to address. Use these notes to
decide what to cover and how to prioritise it. \textbf{Never mention, quote,
restate or allude to the notes themselves} --- the reader must not be able to
tell that you had them. Your output format is unchanged.}}

\smallskip\hrule\smallskip
\textbf{\textcolor{black!70}{[user]}}\\
{\itshape\textcolor{black!45}{(identical to P1, then appended:)}}\\[2pt]
\colorbox{yellow!22}{\parbox{0.965\linewidth}{%
\texttt{\# What this user is actually looking for (private editorial notes):}\\[2pt]
\texttt{\#\# The user's own elaboration of the question:}\\
\texttt{\{details\}}\\[3pt]
\texttt{\#\# The specific aspects this user expects the answer to address:}\\
\texttt{- \{aspect\}}\\
\hspace*{1.5em}\texttt{- why it matters to this user: \{reason\}}\\
\hspace*{1.5em}\texttt{- in the user's own words: \{evidence\}}\\
\hspace*{1.5em}{\scriptsize\itshape(repeated for every aspect in the rubric)}}}
\end{promptbox}

\vspace{4pt}

\begin{promptbox}{P3 — Judge}
\footnotesize
\textbf{\textcolor{black!70}{[system]}}\\
You are a fair and insightful judge with exceptional reasoning and analytical
abilities. Your task is to evaluate a user's question, a generated response to
that question, and an aspect that is important to the user. Based on this
information, identify if the aspect is addressed in the generated response.\\[2pt]
\texttt{\# your output:} [\ldots] a valid json object [\ldots] that contains:\\
\hspace*{1em}- \texttt{match\_score}: A score between 0 to 2 [\ldots] where
\textbf{0} means the response does not cover this aspect, \textbf{1} means the
response somewhat covers this aspect, and \textbf{2} means the response covers
this aspect very well.

\smallskip\hrule\smallskip
\textbf{\textcolor{black!70}{[user]}}\\
\texttt{question: \{question\}}\\
\texttt{details: \{details\}}\\
\texttt{response: \{response\}}\\
\texttt{aspect: \{aspect, reason, evidence\}}
\end{promptbox}

\end{document}